\documentclass[10pt,twocolumn,letterpaper]{article}

\usepackage{iccv}              
\usepackage[accsupp]{axessibility}  
\usepackage{float}  
\usepackage{multirow}
\usepackage{tabularx}
\usepackage{graphicx}
\usepackage{dblfloatfix}  
\usepackage{bm} 
\usepackage{pifont}       

\newcommand{\cmark}{\textcolor{green}{\ding{51}}}  
\newcommand{\xmark}{\textcolor{red}{\ding{55}}}      

\definecolor{gold}{HTML}{FFD700}    
\definecolor{silver}{HTML}{C0C0C0}  
\definecolor{bronze}{HTML}{CD7F32}  

\newcommand{\capscore}[2]{{\setlength{\fboxsep}{1pt}\colorbox{#1}{\textbf{#2}}}}

\definecolor{iccvblue}{rgb}{0.21,0.49,0.74}
\usepackage[pagebackref,breaklinks,colorlinks,allcolors=iccvblue]{hyperref}

\def\paperID{6544} 
\def\confName{ICCV}
\def\confYear{2025}

\title{Towards Open-World Generation of Stereo Images and Unsupervised Matching}

\author{
Feng Qiao \quad
Zhexiao Xiong \quad
Eric Xing  \quad
Nathan Jacobs \\
Washington University in St. Louis
}
\begin{document}
\twocolumn[{%
\renewcommand\twocolumn[1][]{#1}%
\maketitle
\begin{center}
    \centering
    \vspace{-4ex}
    \begin{minipage}[b]{0.02\textwidth}
    \end{minipage}
    \includegraphics[width=\textwidth]{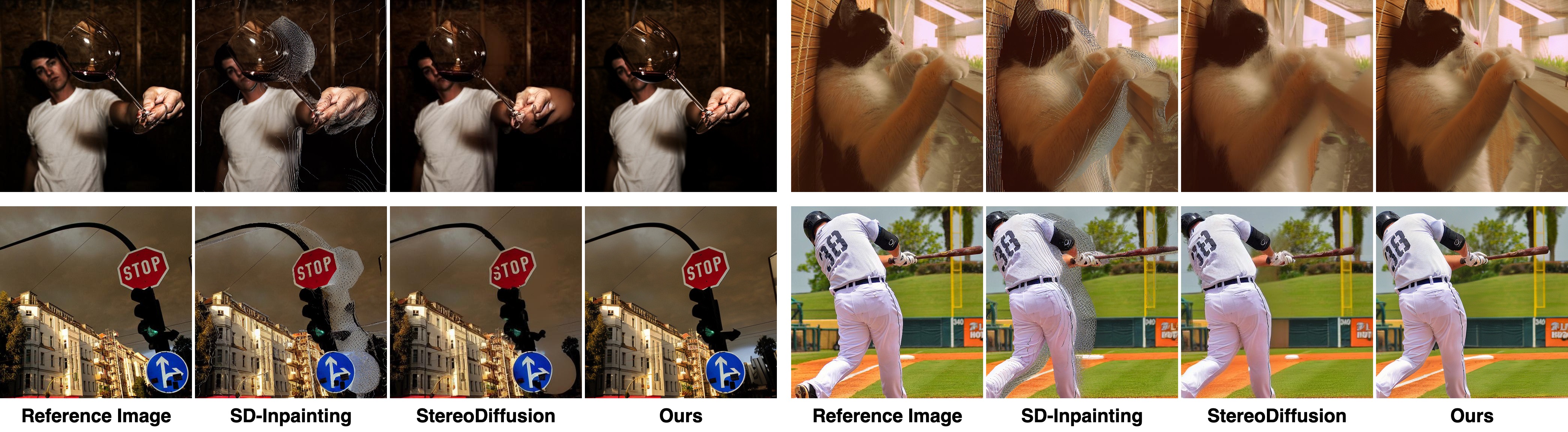}  \\[-2ex]
    \captionof{figure}{Comparison with diffusion-based methods for stereo image generation on the COCO dataset.}    
    \label{fig:teaser}
\end{center}%
}]

\begin{abstract}

    Stereo images are fundamental to numerous applications, including extended
    reality (XR) devices, autonomous driving, and robotics. Unfortunately,
    acquiring high-quality stereo images remains challenging due to the precise
    calibration requirements of dual-camera setups and the complexity of
    obtaining accurate, dense disparity maps. Existing stereo image generation
    methods typically focus on either visual quality for viewing or geometric
    accuracy for matching, but not both. We introduce GenStereo, a diffusion-based
    approach, to bridge this gap. The method includes two primary innovations (1)
    conditioning the diffusion process on a disparity-aware coordinate embedding
    and a warped input image, allowing for more precise stereo alignment than
    previous methods, and (2) an adaptive fusion mechanism that intelligently combines
    the diffusion-generated image with a warped image, improving both realism
    and disparity consistency. Through extensive training on 11 diverse stereo datasets,
    GenStereo demonstrates strong generalization ability. GenStereo achieves state-of-the-art
    performance in both stereo image generation and unsupervised stereo matching
    tasks. Project page is available at https://qjizhi.github.io/genstereo.

\end{abstract}
\section{Introduction}
\label{sec:intro}

The demand for stereo images continues growing with XR devices, autonomous
driving, and robotics. However, acquiring high-quality stereo images remains challenging
due to complex camera calibration and environmental constraints, limiting the
development of robust stereo matching models that generalize well across diverse
scenarios. Real-world datasets often provide only sparse disparity annotations~\cite{menze2015object}
or lack accurate ground truth altogether~\cite{cho2021deep}. Although some
datasets offer dense, accurate disparity maps~\cite{bao2020instereo2k, ramirez2023booster},
they are typically limited to specific scenarios like indoor scenes. Moreover,
capturing real-world stereo data requires complex sensor setups with precise
calibration, often constraining baseline distances and scene diversity.
Synthetic datasets, while providing precise disparity maps, suffer from domain
gaps compared to real-world scenarios~\cite{cabon2020virtual, wang2020tartanair,
wang2021irs}.

Recent advancements in monocular depth estimation (MDE) models~\cite{ranftl2020towards,bhat2023zoedepth,yang2024depth_cvpr,yang2024depth_neurips}
have provided increasingly accurate dense disparity maps, playing a crucial role
in image generation from single views. Meanwhile, Text-to-Image (T2I) diffusion models,
such as Stable Diffusion~\cite{Rombach_2022_CVPR}, have made remarkable strides in
generating diverse and high-quality images from user-provided text prompts.
However, these models often struggle to maintain spatial consistency when altering
the viewpoint of the generated images. While there has been some research on 3D
novel view generation~\cite{tseng2023consistent, Seo2024GenWarpSI}, there remains a
gap in utilizing image generation models to directly produce stereo images.
StereoDiffusion~\cite{wang2024stereodiffusion} is the first approach to leverage
diffusion-based generation for stereo image generation, but it struggles with pixel-level
accuracy, as it applies disparity shifts in the latent space and fills occluded
regions with blurry pixels that lack meaningful semantic content. On the other hand,
Stable Diffusion Inpainting (SD-Inpainting)~\cite{Rombach_2022_CVPR}, though not
designed for stereo image generation, faces similar challenges by incorrectly
filling occluded areas with semantically inappropriate pixels. ~\cref{fig:teaser}
shows that previous diffusion-based methods and our methods tested on the COCO
dataset~\cite{lin2014microsoft}. The scale factor $\gamma$ is set to 0.15,
meaning that the maximum disparity is 15\% of the image width.

We present GenStereo, a novel framework that addresses both visual quality and
geometric accuracy in stereo image generation. Existing approaches rely on either
warping-based methods~\cite{watson2020learning}, which provide geometric
accuracy but struggle with semantic consistency in occluded regions, or diffusion-based
approaches~\cite{wang2024stereodiffusion}, which maintain better semantic
coherence but lack precise geometric control. GenStereo bridges this gap through
a carefully designed two-stream architecture inspired by recent advances in human
animation~\cite{hu2024animate} and view synthesis~\cite{Seo2024GenWarpSI},
achieving both geometric precision and semantic consistency.

Our framework employs a multi-level constraint system for geometric precision
and semantic coherence, starting with disparity-aware coordinate embedding that provides
implicit geometric guidance. This is followed by a cross-view attention
mechanism that enables semantic feature alignment between views, ensuring consistency
in challenging areas like occlusions and complex textures. The framework
leverages a dual-space supervision strategy that operates in both latent and
pixel spaces, with an adaptive fusion mechanism that further optimizes pixel-level
accuracy. To ensure robust generalization across diverse real-world scenarios, we
train our model on a diverse dataset combining 11 stereo datasets with varying scenes
and baselines.

Furthermore, we demonstrate how our high-quality stereo generation framework
significantly improves unsupervised stereo matching learning. Previous approaches
to unsupervised learning have been limited by either simplified warping and random
background filling~\cite{watson2020learning} or constraints to small-scale static
scenes~\cite{tosi2023nerf}. In contrast, our method enables large-scale training
with diverse, photorealistic stereo images that maintain both geometric accuracy
and semantic consistency. This advancement represents a significant step toward bridging
the gap between supervised and unsupervised stereo matching approaches. Our
contributions can be summarized as follows:

\begin{itemize}
    \item We propose GenStereo, the first unified framework for open-world
        stereo image generation that addresses visual quality and geometric
        accuracy, enabling both practical applications and unsupervised stereo matching.

    \item We introduce a comprehensive multi-level constraint system that combines:
        (1) disparity-aware coordinate embedding with warped image conditioning
        for geometric guidance, (2) cross-view attention mechanism for semantic feature
        alignment, and (3) dual-space supervision with adaptive fusion for pixel-accuracy
        generation.

    \item Extensive experimental validation demonstrating state-of-the-art performance
        in both stereo image generation and unsupervised stereo matching.
\end{itemize}
\section{Related Work}

\subsection{Conditional Diffusion Models}
Diffusion models began with DDPM~\cite{ho2020denoising}, advancing from slow, probabilistic
generation to faster sampling with DDIM~\cite{Song2021DenoisingDI}. Stable
Diffusion~\cite{Rombach_2022_CVPR}, a pioneering T2I model, further improved
efficiency by performing diffusion in a compact latent space, combining it with CLIP-based~\cite{radford2021learning}
text conditioning for versatile T2I generation. However, Stable Diffusion lacks fine-grained
control over specific visual attributes, relying mainly on text prompts without
structured guidance. To address these limitations, conditional diffusion models
have emerged. ControlNet~\cite{zhang2023adding} is a key advancement,
introducing conditioning mechanisms for structural inputs (such as edges, depth
maps, or poses) that maintain flexibility while enabling precise, user-driven synthesis.
However, ControlNet still faces challenges in achieving pixel-level accuracy in complex
images, such as stereo image generation.

Recent research~\cite{khachatryan2023text2video, shi2023mvdream, hu2024animate, xu2024magicanimate,
Seo2024GenWarpSI}
has explored the self-attention properties within T2I models. Text2Video-Zero~\cite{khachatryan2023text2video}
and MVDream~\cite{shi2023mvdream} generate consistent visuals across video frames
or 3D multi-views by sharing self-attention, while Animate-Anyone~\cite{hu2024animate}
and MagicAnimate~\cite{xu2024magicanimate} apply a similar approach to produce
human dance videos through fine-tuned T2I models. GenWarp~\cite{Seo2024GenWarpSI} employs
augmenting self-attention with cross-view attention between the reference and target
views to generate novel images. This two-stream architecture augments the
features of the denoising net. Inspired by the adaptability and control of these
self-attention-based architectures, our method builds on their strengths. However,
their direct application to stereo image generation faces fundamental challenges.
First, they typically don't incorporate disparity information, which is crucial for
stereoscopic consistency. Second, achieving the pixel-level accuracy required for
comfortable stereo viewing demands specialized architectural considerations beyond
existing frameworks.

\subsection{Stereo Image Generation and Inpainting}
Traditional view synthesis approaches, including geometry-based reconstruction~\cite{hedman2017casual,
hedman2018deep, kopf2013image}
and recent NeRF-based methods~\cite{tosi2023nerf}, excel at rendering novel views
but are constrained by their requirement for multiple input images of static scenes.
While 3D Photography techniques~\cite{hedman2017casual, shih20203d} attempt to
overcome this limitation through depth-based mesh projection, they often struggle
with complex occlusion handling and require sophisticated post-processing pipelines.

The evolution of stereo generation methods has progressed from simple warping-based
approaches to more sophisticated diffusion-based solutions. MfS~\cite{watson2020learning}
introduced a basic framework that attempted to handle occlusions by sampling
random background patches from the dataset, but this led to semantic inconsistencies
and visible artifacts at occlusion areas. While SD-Inpainting~\cite{Rombach_2022_CVPR}
introduced more coherent semantic content through its prior learning, it failed to
maintain local consistency between inpainted regions and their surrounding context,
often producing visible discontinuities in texture and structure. Mono2Stereo~\cite{wang2024mono2stereo}
made progress by fine-tuning SD-Inpainting specifically for stereo generation,
enabling unsupervised stereo matching, though visible artifacts persist in in-painted regions. StereoDiffusion~\cite{wang2024stereodiffusion} marked a significant
shift by introducing end-to-end diffusion-based stereo image generation, but its
latent-space warping approach without explicit geometric constraints led to compromised
pixel-level accuracy. Despite these advances, the challenge of generating photorealistic
stereo images that maintain both visual quality and geometric fidelity remains largely
unsolved, particularly for diverse real-world scenarios.

\subsection{Unsupervised Stereo Matching}
Traditional unsupervised stereo matching approaches primarily rely on
photometric consistency. Methods like \cite{godard2019digging, tonioni2019learning,
tonioni2019real, zhong2017self} leverage photometric losses across stereo images,
while others \cite{chi2021feature, lai2019bridging, wang2019unos} extend this to
temporal sequences. A parallel line of research explores proxy supervision strategies,
either through carefully designed algorithmic supervisors
\cite{poggi2021continual, tonioni2017unsupervised, tonioni2019unsupervised} or
knowledge distillation from pre-trained networks \cite{aleotti2020reversing}. Recent
domain adaptation approaches \cite{song2021adastereo, liu2020stereogan,
Xiong_2023_BMVC} attempt to bridge the gap between synthetic and real-world
domains or facilitate cross-dataset adaptation. However, these methods often exhibit
limited generalization capability beyond their target domains \cite{aleotti2020reversing},
particularly struggling with diverse real-world scenarios.

More recent approaches have explored novel directions for stereo generation and
matching. MfS~\cite{watson2020learning} pioneered unsupervised stereo matching using
MDE as guidance, but it relies on random background sampling for occlusion
filling, leading to semantic inconsistencies. While NeRF-Stereo~\cite{tosi2023nerf}
achieves high-quality results through neural radiance fields, it requires multiple
views of static scenes, limiting its practical applications. Mono2Stereo~\cite{wang2024mono2stereo}
leverages Stable Diffusion for inpainting occluded regions, representing a
significant advance in generative stereo synthesis, though it still exhibits artifacts
in occluded areas where geometric and semantic consistency is crucial.
\section{Methods}
\begin{figure*}[h]
    \centering
    \includegraphics[width=\linewidth]{
        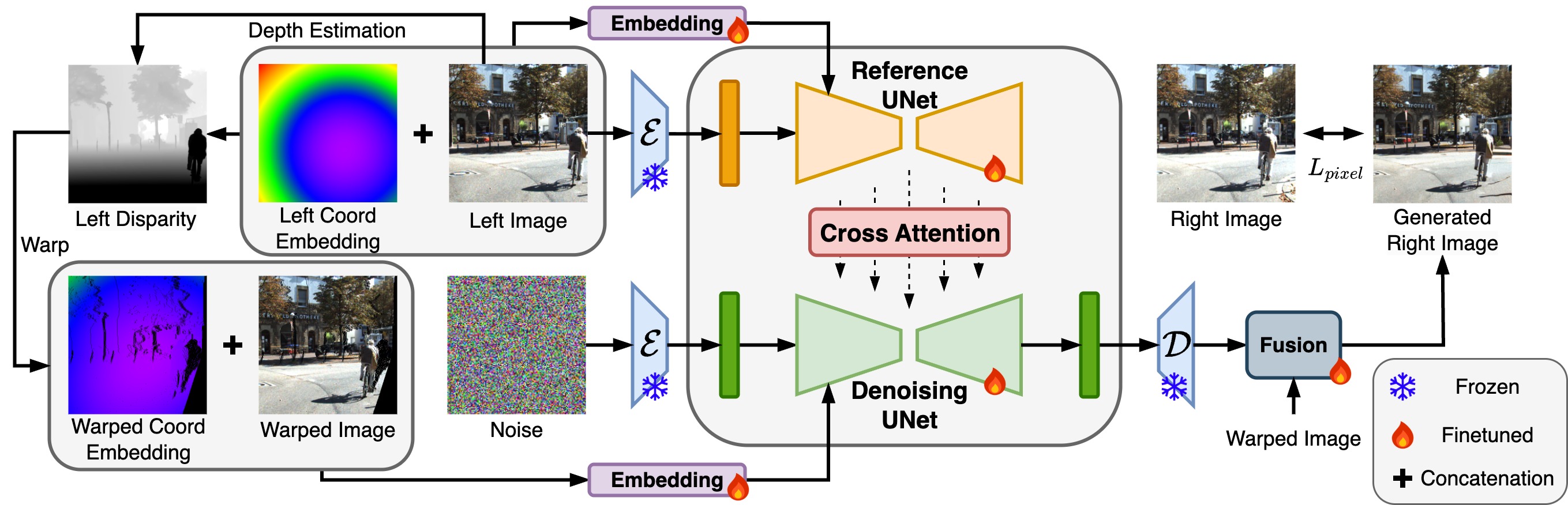
    } 
    \caption{Overview of the GenStereo framework.}
    \label{fig:framework}
\end{figure*}

Given a left view image $I_{l}$ and its corresponding disparity map
$\hat{D}_{l}$ predicted by an MDE model, our goal is to generate a high-quality right
view image $\hat{I}_{r}$ that maintains both visual quality and geometric consistency.
As shown in ~\cref{fig:framework}, our GenStereo framework processes the left
view as the reference image and synthesizes the right view as the target image. The
generated triplet $\langle I_{l}, \hat{D}_{l}, \hat{I}_{r}\rangle$ not only serves
as a stereo pair for visualization but also provides training data for unsupervised
stereo matching.

\subsection{Disparity-Aware Coordinate Embedding}
Traditional inpainting-based stereo image generation methods often suffer from
visible boundaries between warped and inpainted regions. To address this limitation,
we propose a disparity-aware coordinate embedding scheme inspired by recent
advances in coordinate-based generation~\cite{mu2022coordgan, Seo2024GenWarpSI}. Our
approach utilizes dual coordinate embeddings: a canonical embedding for the left
view and its warped counterpart for the right view.

Specifically, we first construct a canonical 2D coordinate map $X \in \mathbb{R}^{h
\times w \times 2}$ with values normalized to $[-1, 1]$. This map is transformed
into Fourier features~\cite{tancik2020fourier} through a positional encoding
function $\phi$:
\begin{equation}
    \begin{aligned}
        C_{l} & = \phi(X)
    \end{aligned}
\end{equation}
The resulting Fourier feature map $C_{l}$ serves as the coordinate embedding for
the left view $I_{l}$. We then generate the right view embedding by warping
$C_{l}$ according to the disparity map:
\begin{equation}
    \begin{aligned}
        C_{r}= \text{warp}(C_{l}, D_{l}),
    \end{aligned}
\end{equation}
where $D_{l}$ represents ground-truth disparity during training and predicted disparity
$\hat{D}_{l}$ during inference. These coordinate embeddings ($C_{l}$ and $C_{r}$)
are integrated into their respective view features ($F_{l}$ and $F_{r}$) through
convolutional layers, establishing strong geometric correspondence while maintaining
visual consistency between views.

Our approach differs from GenWarp~\cite{Seo2024GenWarpSI} by utilizing disparity maps
instead of camera matrices for warping, enabling more precise pixel-level
control. We further enhance geometric consistency by incorporating the warped image
$I_{\text{warp}}$ as additional conditioning for the denoising U-Net.

\subsection{Cross-View Feature Enhancement}
To facilitate effective information exchange between views, we adopt a two-parallel
U-Nets framework pretrained from Stable Diffusion, inspired by GenWarp~\cite{Seo2024GenWarpSI}.
However, our work differs in its novel conditioning strategy: the reference U-Net
takes concatenated $(I_{l}, C_{l})$ as conditions to process the left view, while
the denoising U-Net takes concatenated $(I_{\text{warp}}, C_{r})$ as conditions when
synthesizing the right view $\hat{I}_{r}$. This design ensures that each U-Net
receives both image content and corresponding coordinate information for better
feature extraction. As our model is conditioned on a reference image and its disparity
map, we replace text condition of SD with image embedding of the left image
through a CLIP image encoder in both U-Nets.

To incorporate left-view information during right-view generation, we concatenate
the reference features with the target features in the attention mechanism.
Specifically, we compute cross-view attention as:
\begin{equation}
    q = F_{r}, \quad k = [F_{l}, F_{r}], \quad v = [F_{l}, F_{r}].
\end{equation}
where $F_{l}$ is derived from the reference U-Net conditioned on
$(I_{l}, C_{l})$, while $F_{r}$ comes from the denoising U-Net conditioned on $(I
_{\text{warp}}, C_{r})$. This dual-stream attention mechanism allows the model
to adaptively balance between semantic consistency from the reference view and
geometric accuracy from the warped view, with each stream guided by both image content
and coordinate information.

\subsection{Training Strategy}
Mixed training~\cite{yang2024depth_cvpr, ranftl2020towards, li2022practical} has
been proven to be an effective training strategy for domain generalization in both
monocular and stereo matching depth estimation. We follow the datasets
summarized in ~\cite{Tosi_IJCV_2025} and utilize the most widely available public
synthetic datasets. Based on our observations, real-world datasets can negatively
impact performance, even when there are only minor calibration errors or slight
differences in imaging between the two cameras. ~\cref{tab:datasets} shows the
comprehensive list of datasets utilized. Datasets containing abstract images are
not included, because our model aims to have a good generalization capability in
the real world.

To address dataset imbalance, we employ a resampling strategy. Specifically, we sample
the smaller datasets multiple times until the number reaches $10\%$ of the largest
dataset. As the Stable Diffusion model only accepts images of size
$512 \times 512$ for SD v1.5 and $768 \times 768$ for SD v2.1, we apply a random
square crop and resize.

\begin{table}[t!]
    \centering
    \scriptsize
    \caption{Our training data sources. All datasets are synthetic with the size
    of 684K.}
    \begin{tabular}{l|c|c|c|c}
        \toprule \textbf{Dataset}                      & \textbf{Indoor} & \textbf{Outdoor} & \textbf{\#Images} & \textbf{Year} \\
        \midrule TartanAir~\cite{wang2020tartanair}    & \cmark          & \cmark           & 306K              & 2020          \\
        Dynamic Replica~\cite{karaev2023dynamicstereo} & \cmark          &                  & 145K              & 2023          \\
        IRS~\cite{wang2021irs}                         & \cmark          &                  & 103K              & 2021          \\
        Falling Things~\cite{tremblay2018falling}      & \cmark          & \cmark           & 61K               & 2018          \\
        VKITTI2~\cite{cabon2020virtual}                &                 & \cmark           & 21K               & 2020          \\
        InfinigenSV~\cite{jing2024match}               &                 & \cmark           & 17K               & 2024          \\
        SimStereo~\cite{gf1e-t452-22}                  & \cmark          &                  & 14K               & 2022          \\
        UnrealStereo4K~\cite{tosi2021smd}              & \cmark          & \cmark           & 8K                & 2021          \\
        Spring~\cite{Mehl2023_Spring}                  &                 & \cmark           & 5K                & 2023          \\
        PLT-D3~\cite{DVN/36SQKM_2024}                  &                 & \cmark           & 3K                & 2024          \\
        Sintel~\cite{Butler:ECCV:2012}                 &                 & \cmark           & 1K                & 2012          \\
        \bottomrule
    \end{tabular}
    \label{tab:datasets}
\end{table}

\subsection{Pixel Space Alignment}
Given an RGB image $I \in \mathbb{R}^{H \times W \times 3}$, Latent Diffusion Models
(LDMs) first encode it into a latent representation $z = \mathcal{E}(I)$, where
$z \in \mathbb{R}^{h \times w \times c}$. While this latent space operation significantly
reduces computational costs, it may compromise pixel-level accuracy during image
generation.

The standard LDM objective operates in latent space:
\begin{equation}
    \begin{aligned}
        L_{latent}:= \mathbb{E}_{z, \epsilon, t}\left[ \| \epsilon - \epsilon_{\theta}(z_{t}, t) \|_{2}^{2}\right]
    \end{aligned}
\end{equation}
where $z_{t}$ is the noisy latent at timestep $t$, $\epsilon$ is the noise added
to create $z_{t}$, and $\epsilon_{\theta}(z_{t}, t)$ is the network's prediction
of this noise. The model learns to denoise by predicting the noise that was
added.

To maintain pixel-level accuracy, we introduce an additional pixel-space loss by
decoding both the predicted and target latent variables:
\begin{equation}
    \begin{aligned}
        L_{pixel}= \mathbb{E}_{z, \epsilon, t}\left[ \| \mathcal{D}(z_{\text{pred}}) - \mathcal{D}(z_{\text{target}}) \|_{2}^{2}\right]
    \end{aligned}
\end{equation}
The final objective combines both spaces:
\begin{equation}
    \begin{aligned}
        L = L_{latent}+ \alpha L_{pixel}
    \end{aligned}
\end{equation}
where $\alpha = 1$ balances latent and pixel-space supervision.

\subsection{Adaptive Fusion Module}
To achieve seamless integration between generated and warped content, we propose
an Adaptive Fusion Module that learns to combine $I_{gen}$ and $I_{warp}$ based
on local context and confidence. The module predicts spatially-varying fusion weights
through a lightweight network:
\begin{equation}
    W = \sigma\left(f_{\theta}\left(\text{concat}(I_{gen}, I_{warp}, M) \right) \right
    ),
\end{equation}
where $f_{\theta}$ is a $3 \times 3$ convolutional layer and $\sigma$ is the
sigmoid activation, ensuring weights in $[0,1]$. The final right view is
computed as:
\begin{equation}
    \hat{I}_{r}= M \odot W \odot I_{warp}+ (1 - M \odot W) \odot I_{gen},
\end{equation}
where $\odot$ denotes element-wise multiplication. This formulation adaptively
favors warped content in high-confidence regions ($M \approx 1$) while relying
on generated content in occluded or uncertain areas. The learned weights $W$ enable
smooth transitions between warped and generated regions, ensuring both geometric
accuracy and visual consistency.

\subsection{Random Disparity Dropout}
To simulate the sparsity of real-world disparity maps like KITTI's LiDAR-derived
ground truth, we randomly apply disparity dropout to 10\% of training samples. For
each selected sample, we first generate a dropout ratio:
\begin{equation}
    r \sim \text{Uniform}(0, 1)
\end{equation}

Using this ratio, we create a binary mask where each pixel has a probability $r$
of being dropped:
\begin{equation}
    M_{\text{rand}}(i,j) =
    \begin{cases}
        1 & \text{otherwise}           \\
        0 & \text{with probability } r
    \end{cases}
\end{equation}
The final mask combines this random dropout with the warping mask:
\begin{equation}
    M = M_{\text{warp}}\lor M_{\text{rand}}
\end{equation}
where $\lor$ denotes element-wise logical OR. This strategy encourages the model
to handle sparse disparity inputs, improving its robustness and generalization to
real-world scenarios where dense disparity maps may not be available.
\section{Experiments}

\subsection{Stereo Image Generation}

\noindent
\textbf{Experimental Setup.} We fine-tune the pretrained Stable Diffusion (SD) UNet for
3 epochs across all experiments. For evaluation, we
utilize two widely adopted stereo vision benchmarks: Middlebury 2014~\cite{scharstein2014high}
and KITTI 2015~\cite{menze2015object}, neither of which are included in our
training set. Middlebury 2014 comprises 23 high-resolution indoor stereo pairs captured
with wide-baseline stereo cameras under varying illumination conditions,
providing challenging scenarios for stereo generation. KITTI 2015 contains 200 outdoor
stereo pairs of street scenes with LiDAR-based sparse disparity ground truth,
offering diverse real-world testing scenarios. Following StereoDiffusion~\cite{wang2024stereodiffusion},
we preprocess Middlebury images to $512 \times 512$ resolution, while KITTI images
are center-cropped before resizing to $512 \times 512$ to maintain the most informative
regions.

We conduct comprehensive comparisons against both traditional and learning-based
baselines. Traditional approaches include na\"ive solutions such as leave blank
and image stretching. For learning-based comparisons, we evaluate against state-of-the-art
diffusion models adapted for stereo generation: Repaint~\cite{lugmayr2022repaint}
and SD-Inpainting~\cite{Rombach_2022_CVPR}, as well as StereoDiffusion, which specifically
targets stereo image generation. To quantitatively assess the quality of
generated right-view images ($\hat{I_r}$) against ground truth ($I_{r}$), we employ
three complementary metrics: Peak Signal-to-Noise Ratio (PSNR), Structural Similarity
Index Measure (SSIM), and Learned Perceptual Image Patch Similarity (LPIPS) with
SqueezeNet backbone.

\noindent
\textbf{Quantitative Results.} \cref{tab:quantitative_results} presents the generation
performance on Middlebury 2014 and KITTI 2015. The sparsity of ground truth disparities
in KITTI limits evaluation accuracy, making pseudo-label disparities more
effective for assessing stereo consistency. This sparsity particularly affects inpainting-based
methods, as they struggle to infer accurate right-view images in areas with
missing depth information, especially for dynamic objects and distant regions.
In contrast, Middlebury 2014, which provides dense ground truth disparities, demonstrates
a smaller performance gap between using ground truth and using pseudo disparities,
underscoring the advantage of high-quality disparity annotations for generation.
We obtain pseudo disparities from CREStereo~\cite{li2022practical} pretrained model,
where evaluation datasets are not used for training.


\begin{table*}
  [th]
  \centering
  \scriptsize
  \caption{Quantitative results of image generation for Middlebury 2014 and
  KITTI 2015 datasets. The top three results for each metric are highlighted with
  a \capscore{gold}{first}, \capscore{silver}{second}, and \capscore{bronze}{third}
  background, respectively.}
  \begin{tabular}{l|c|c c c|c c c}
    \toprule \multirow{2.5}{*}{\textbf{Methods}}   & \multirow{2.5}{*}{\textbf{SD version}} & \multicolumn{3}{c|}{\textbf{Middlebury 2014}} & \multicolumn{3}{c}{\textbf{KITTI 2015}}       \\
    \cmidrule(lr){3-5} \cmidrule(lr){6-8}          &                                        & \textbf{PSNR $\uparrow$}                      & \textbf{SSIM $\uparrow$}                     & \textbf{LPIPS $\downarrow$}                  & \textbf{PSNR $\uparrow$}                      & \textbf{SSIM $\uparrow$}                     & \textbf{LPIPS $\downarrow$}                  \\
    \midrule Leave blank                           & -                                      & $11.328_{-3.489}^{+2.483}$                    & $0.315_{-0.154}^{+0.230}$                    & $0.450_{-0.089}^{+0.097}$                    & $12.980_{-3.831}^{+4.919}$                    & $0.374_{-0.251}^{+0.286}$                    & $0.313_{-0.109}^{+0.222}$                    \\
    Stretch                                        & -                                      & $14.842_{-2.753}^{+2.714}$                    & $0.432_{+0.265}^{-0.190}$                    & $0.285_{-0.089}^{+0.112}$                    & $14.757_{-4.694}^{+5.375}$                    & $0.429_{-0.271}^{+0.287}$                    & $0.212_{-0.100}^{+0.145}$                    \\
    3D Photography~\cite{shih20203d}               & -                                      & $14.190_{-2.798}^{+2.464}$                    & $0.427_{-0.175}^{+0.238}$                    & $0.275_{-0.065}^{+0.073}$                    & $14.540_{-4.023}^{+7.256}$                    & $0.398_{-0.323}^{+0.270}$                    & $0.210_{-0.073}^{+0.099}$                    \\
    RePaint~\cite{lugmayr2022repaint}              & 1.5                                    & $15.102_{-2.802}^{+2.909}$                    & $0.462_{-0.253}^{+0.268}$                    & $0.311_{-0.079}^{+0.088}$                    & $15.056_{-4.897}^{+5.366}$                    & $0.462_{-0.285}^{+0.268}$                    & $0.251_{-0.095}^{+0.128}$                    \\
    SD-Inpainting~\cite{Rombach_2022_CVPR}         & 1.5                                    & $15.740_{-3.351}^{+3.511}$                    & $0.412_{-0.188}^{+0.236}$                    & $0.311_{-0.164}^{+0.098}$                    & $9.792_{-4.070}^{+2.809}$                     & $0.230_{-0.229}^{+0.236}$                    & $0.652_{-0.139}^{+0.144}$                    \\
    StereoDiffusion~\cite{wang2024stereodiffusion} & 1.5                                    & $15.456_{-3.313}^{+2.669}$                    & $0.468_{-0.205}^{+0.252}$                    & $0.231_{-0.088}^{+0.096}$                    & $15.679_{-5.487}^{+5.888}$                    & $0.481_{-0.310}^{+0.245}$                    & $0.205_{-0.099}^{+0.135}$                    \\
    Ours + GT                                      & 1.5                                    & $23.699_{-6.033}^{+5.456}$                    & $0.866_{-0.141}^{+0.084}$                    & $0.064_{-0.034}^{+0.053}$                    & $\capscore{bronze}{20.439}_{-8.541}^{+7.518}$ & $0.746_{-0.274}^{+0.155}$                    & $\capscore{silver}{0.108}_{-0.048}^{+0.117}$ \\
    Ours + Pseudo                                  & 1.5                                    & $\capscore{bronze}{23.835}_{-6.741}^{+6.842}$ & $\capscore{bronze}{0.868}_{-0.157}^{+0.096}$ & $\capscore{bronze}{0.062}_{-0.038}^{+0.063}$ & $\capscore{silver}{22.749}_{-6.102}^{+6.722}$ & $\capscore{silver}{0.813}_{-0.152}^{+0.110}$ & $\capscore{gold}{0.096}_{-0.040}^{+0.068}$   \\
    Ours + GT                                      & 2.1                                    & $\capscore{silver}{24.819}_{-4.252}^{+4.519}$ & $\capscore{silver}{0.906}_{-0.132}^{+0.057}$ & $\capscore{silver}{0.061}_{-0.032}^{+0.059}$ & $19.836_{-8.483}^{+8.952}$                    & $\capscore{bronze}{0.765}_{-0.245}^{+0.137}$ & $0.135_{-0.059}^{+0.104}$                    \\
    Ours + Pseudo                                  & 2.1                                    & $\capscore{gold}{25.142}_{-4.872}^{+6.970}$   & $\capscore{gold}{0.911}_{-0.133}^{+0.066}$   & $\capscore{gold}{0.060}_{-0.037}^{+0.061}$   & $\capscore{gold}{23.488}_{-6.444}^{+7.090}$   & $\capscore{gold}{0.849}_{-0.130}^{+0.092}$   & $\capscore{bronze}{0.109}_{-0.050}^{+0.068}$ \\
    \bottomrule
  \end{tabular}
  \label{tab:quantitative_results}
\end{table*}

\begin{figure*}[th]
  \centering
  \includegraphics[width=\linewidth]{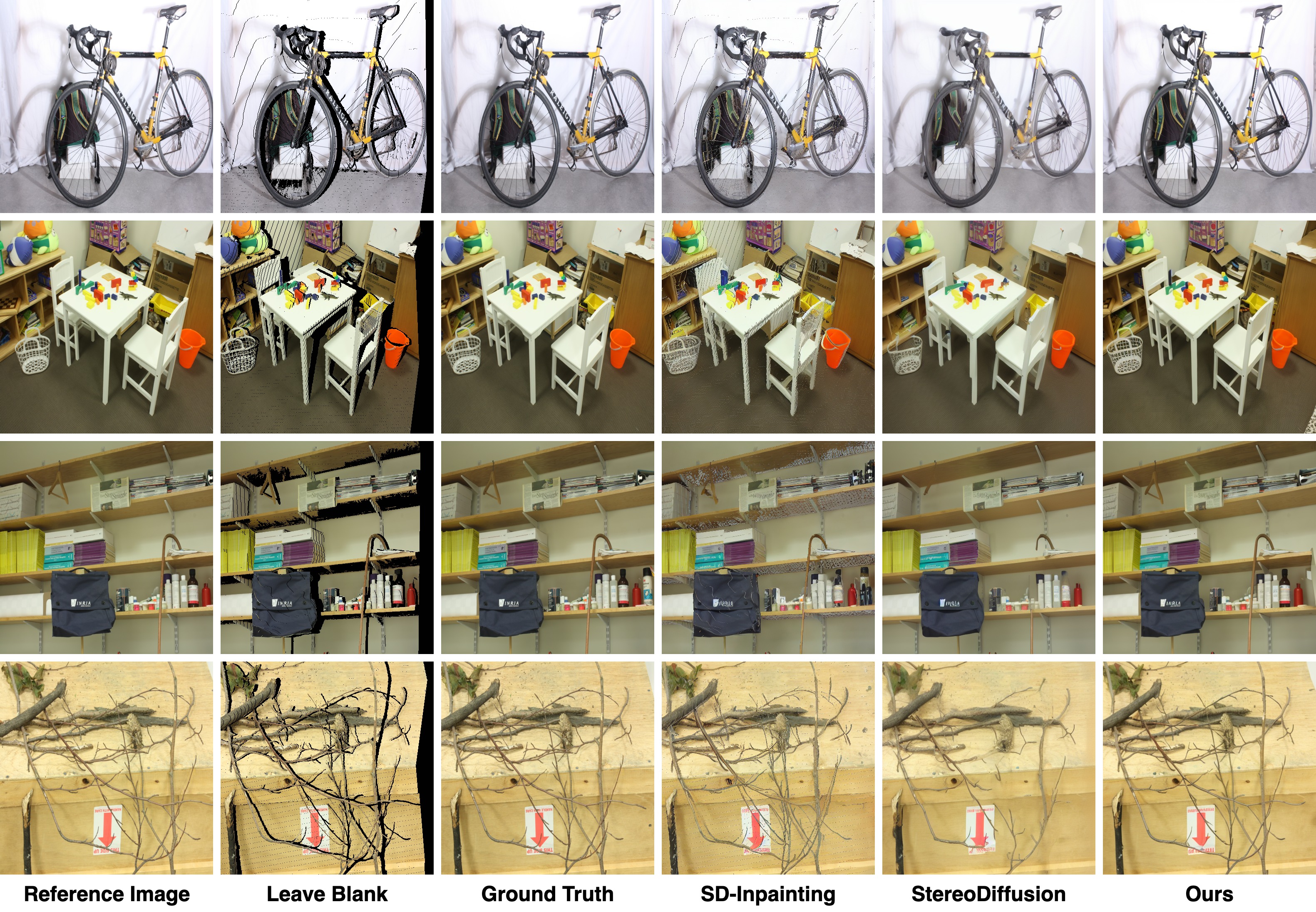}
  \caption{Qualitative comparison on Middlebury 2014 with ground truth disparity
  maps.}
  \label{fig:vis_middlebury}
\end{figure*}

\noindent
\textbf{Qualitative Results.} We present qualitative comparisons between our method
and several baselines: leave blank, StereoDiffusion, and SD-Inpainting, with ground
truth as reference. As shown in \cref{fig:vis_middlebury}, our method demonstrates
superior performance on the Middlebury 2014 dataset. While SD-Inpainting
struggles with occlusion handling, often generating semantically inconsistent content
in occluded regions, and StereoDiffusion exhibits loss of fine details due to
latent-space warping, our approach successfully maintains both geometric accuracy
and visual fidelity. Particularly noteworthy is our method's ability to generate
coherent right-view images while preserving pixel-level correspondence with the
left images. Additional qualitative results on the KITTI 2015 dataset and more visualizations
on other datasets can be found in the supplementary materials.

\begin{table}[ht!]
  \centering
  \scriptsize
  \caption{Comparison with other diffusion-based stereo image generation methods
  for unsupervised stereo matching. The first group of methods is trained on PSMNet
  while the second group is trained on RAFT-Stereo. SD v1.5 is used for the experiments.}
  \begin{tabular}{l|cc|cc}
    \toprule \multirow{2.5}{*}{\textbf{Methods}}   & \multicolumn{2}{c|}{\textbf{KITTI 2012}} & \multicolumn{2}{c}{\textbf{KITTI 2015}} \\
    \cmidrule(lr){2-3} \cmidrule(lr){4-5}          & \textbf{D1-all $\downarrow$}             & \textbf{EPE $\downarrow$}              & \textbf{D1-all $\downarrow$} & \textbf{EPE $\downarrow$} \\
    \midrule                                        
    Leave blank                                    & 25.219                                   & 3.542                                  & 24.568                       & 2.937                     \\
    SD-Inpainting~\cite{Rombach_2022_CVPR}         & 3.907                                    & 0.894                                  & 4.490                        & 1.059                     \\
    StereoDiffusion~\cite{wang2024stereodiffusion} & 15.213                                   & 2.220                                  & 5.651                        & 1.154                     \\
    Ours                                           & \capscore{gold}{3.802}                   & \capscore{gold}{0.815}                 & \capscore{gold}{3.933}       & \capscore{gold}{0.991}    \\
    \midrule                                        
    Leave blank                                    & 4.184                                    & 0.952                                  & 4.074                        & 0.957                     \\
    SD-Inpainting~\cite{Rombach_2022_CVPR}         & 4.016                                    & 0.945                                  & 4.128                        & 0.972                     \\
    StereoDiffusion~\cite{wang2024stereodiffusion} & 12.756                                   & 1.811                                  & 6.360                        & 1.174                     \\
    Ours                                           & \capscore{gold}{3.537}                   & \capscore{gold}{0.930}                 & \capscore{gold}{3.653}       & \capscore{gold}{0.943}    \\
    \bottomrule
  \end{tabular}
  \label{tab:unsupervised_finetune}
\end{table}

\begin{table*}
  [th]
  \centering
  \scriptsize
  \caption{Zero-Shot Generalization Benchmark. The first group shows methods trained
  with PSMNet, the second group shows methods trained with RAFT-Stereo, and the third
  group represents IGEV++.}
  \begin{tabular}{l|c c c|c c c|c c c|c c c|c c c}
    \toprule \multirow{2.5}{*}{\textbf{Methods}}                                                        & \multicolumn{3}{c|}{\textbf{KITTI 2012}} & \multicolumn{3}{c|}{\textbf{KITTI 2015}} & \multicolumn{3}{c}{\textbf{ETH3D}} & \multicolumn{3}{|c}{\textbf{Middlebury-Q}} & \multicolumn{3}{|c}{\textbf{Middlebury-H}} \\
    \cmidrule(lr){2-4} \cmidrule(lr){5-7} \cmidrule(lr){8-10} \cmidrule(lr){11-13} \cmidrule(lr){14-16} & \textbf{D1-all}                          & \textbf{EPE}                             & \textbf{\textgreater3px}           & \textbf{D1-all}                            & \textbf{EPE}                              & \textbf{\textgreater3px} & \textbf{D1-all}          & \textbf{EPE}             & \textbf{\textgreater1px} & \textbf{D1-all}          & \textbf{EPE}             & \textbf{\textgreater2px} & \textbf{D1-all}          & \textbf{EPE}             & \textbf{\textgreater2px} \\
    \midrule PSMNet~\cite{chang2018pyramid}                                                             & 26.69                                    & 3.877                                    & 27.35                              & 28.24                                      & 4.003                                     & 28.53                    & 4.258                    & 1.989                    & 15.74                    & 15.00                    & 4.941                    & 20.20                    & 21.34                    & 9.899                    & 33.26                    \\
    MfS~\cite{watson2020learning}                                                                       & 4.322                                    & 1.009                                    & 4.703                              & 4.963                                      & 1.636                                     & 5.181                    & 2.204                    & 0.535                    & 8.170                    & 9.217                    & 1.648                    & 12.07                    & 12.34                    & 3.242                    & 17.56                    \\
    MfS+DAMv2~\cite{yang2024depth_neurips}                                                              & 4.931                                    & 1.063                                    & 5.313                              & 5.504                                      & 1.401                                     & 5.737                    & 1.990                    & 0.493                    & 8.145                    & 9.447                    & 1.825                    & 11.90                    & 12.34                    & 3.678                    & 17.21                    \\
    Mono2Stereo~\cite{wang2024mono2stereo}                                                              & 4.077                                    & 0.859                                    & -                                  & 4.507                                      & 1.052                                     & -                        & 1.531                    & 0.397                    & -                        & -                        & -                        & -                        & -                        & -                        & -                        \\
    NeRFStereo~\cite{tosi2023nerf}                                                                      & 3.683                                    & 0.851                                    & 4.507                              & 4.813                                      & 1.448                                     & 5.041                    & 1.620                    & 0.557                    & 11.68                    & 7.619                    & 1.451                    & 11.04                    & 8.892                    & 2.590                    & 12.88                    \\
    Ours + SD v1.5                                                                                      & 3.676                                    & 0.846                                    & 4.063                              & \capscore{silver}{3.879}                   & 0.996                                     & \capscore{bronze}{4.112} & 1.114                    & 0.378                    & 5.556                    & 6.887                    & 1.139                    & 10.43                    & 11.33                    & 2.269                    & 19.39                    \\
    Ours + SD v2.1                                                                                      & 3.807                                    & 0.820                                    & 4.122                              & \capscore{bronze}{3.898}                   & \capscore{bronze}{0.978}                  & \capscore{silver}{4.096} & 0.894                    & 0.324                    & 4.101                    & 5.037                    & \capscore{bronze}{0.860} & \capscore{bronze}{7.463} & 8.046                    & 1.709                    & 13.73                    \\
    \midrule RAFT-Stereo~\cite{lipson2021raft}                                                          & 3.944                                    & 0.830                                    & 4.239                              & 5.254                                      & 1.132                                     & 5.468                    & \capscore{bronze}{0.883} & \capscore{gold}{0.267}   & \capscore{gold}{2.607}   & 5.157                    & 1.217                    & 10.25                    & 5.774                    & \capscore{silver}{1.436} & 11.20                    \\
    NeRFStereo~\cite{tosi2023nerf}                                                                      & 3.787                                    & 0.855                                    & 4.140                              & 5.356                                      & 1.469                                     & 5.561                    & \capscore{gold}{0.830}   & 0.301                    & 3.370                    & 5.819                    & 1.003                    & 8.370                    & 7.426                    & 1.820                    & 10.95                    \\
    Ours + SD v1.5                                                                                      & \capscore{bronze}{3.224}                 & \capscore{bronze}{0.734}                 & \capscore{bronze}{3.455}           & 4.683                                      & 1.042                                     & 4.897                    & 0.959                    & \capscore{bronze}{0.294} & \capscore{bronze}{2.667} & \capscore{gold}{3.450}   & 1.003                    & 7.952                    & \capscore{silver}{4.446} & \capscore{bronze}{1.514} & \capscore{bronze}{9.94}  \\
    Ours + SD v2.1                                                                                      & \capscore{silver}{3.219}                 & \capscore{silver}{0.716}                 & \capscore{silver}{3.019}           & 3.950                                      & \capscore{silver}{0.971}                  & 4.222                    & \capscore{silver}{0.835} & \capscore{silver}{0.277} & \capscore{silver}{2.638} & \capscore{bronze}{4.067} & 1.127                    & 9.382                    & \capscore{bronze}{4.618} & 1.527                    & 11.11                    \\
    \midrule IGEV++~\cite{xu2025igev++}                                                                 & 5.681                                    & 1.099                                    & 6.196                              & 6.028                                      & 1.265                                     & 6.239                    & 1.701                    & 0.356                    & 4.666                    & 4.238                    & \capscore{silver}{0.828} & \capscore{silver}{6.274} & 5.292                    & 2.384                    & \capscore{silver}{7.752} \\
    Ours + SD v2.1                                                                                      & \capscore{gold}{2.802}                   & \capscore{gold}{0.686}                   & \capscore{gold}{3.015}             & \capscore{gold}{3.746}                     & \capscore{gold}{0.952}                    & \capscore{gold}{4.004}   & 1.226                    & 0.326                    & 3.429                    & \capscore{silver}{3.682} & \capscore{gold}{0.684}   & \capscore{gold}{5.512}   & \capscore{gold}{2.797}   & \capscore{gold}{0.842}   & \capscore{gold}{5.098}   \\
    \bottomrule
  \end{tabular}
  \label{tab:unsupervised_generalize}
\end{table*}

\subsection{Unsupervised Learning}
\noindent
\textbf{Experimental Setup.} We conduct two comprehensive experiments to validate
the effectiveness of our proposed generation method for unsupervised stereo matching.

In the first experiment, we fine-tune a stereo matching model pretrained on the
SceneFlow dataset using the KITTI 2012 and KITTI 2015 benchmarks. We split the training
and evaluation sets into 160/34 images for KITTI 2012 and 160/40 images for KITTI
2015, following~\cite{watson2020learning}.

The second experiment evaluates various unsupervised stereo matching approaches
by generating stereo images. Specifically, KITTI 2012~\cite{geiger2012we} has
194 stereo images, KITTI 2015 has 200 stereo images, ETH3D~\cite{schops2017multi}
has 27 stereo pairs, and the Middlebury v3 training set~\cite{scharstein2014high}
has 15 stereo pairs at Quarter and Half resolutions (Midd-Q, Midd-H). This diverse
dataset selection allows us to comprehensively assess the generalization capability
of our method across different domains and resolutions.

We evaluate the performance of various unsupervised stereo matching approaches using
the D1-all metric, which measures the percentage of incorrect disparity
predictions (error $> 3$px or $5\%$ of the true disparity), the End-Point-Error
(EPE), and the $\textgreater n$px metric, which denotes the percentage of pixels
with an EPE exceeding $n$ pixels. To demonstrate the architecture-agnostic
nature of our approach, we employ three distinct stereo models: PSMNet~\cite{chang2018pyramid},
a cost-volume-based method, RAFT-Stereo~\cite{lipson2021raft}, an iterative refinement
model, and IGEV++~\cite{xu2025igev++}, one of the most recent methods. For disparity estimation, we utilize the Depth Anything Model v2 (DAMv2)~\cite{yang2024depth_neurips},
from which we obtain the disparity map. The disparity values are normalized to the
range $[0 , 1]$ and subsequently scaled by a factor $\gamma$ to enable flexibility
generation.

\noindent
\textbf{Fine-tuning on KITTI.} To ensure fair and consistent comparisons across
all generation methods, we construct the training datasets using a standardized procedure.
Specifically, we sample the scale factor $\gamma$ from the set $\{0.05, 0.1, 0.15
, 0.2, 0.25\}$. For each selected scale, we employ various generation methods to
create the right images based on 160 left images and their corresponding scaled disparity
maps for both KITTI 2012 and KITTI 2015. This procedure results in a total of 800
stereo pairs.

As shown in \cref{tab:unsupervised_finetune}, higher generation quality leads to
improved stereo matching performance. Among the compared methods, SD-Inpainting
achieves the closest performance to ours due to its pixel-level warping
operation. However, it falls short in providing semantic consistency in occluded
areas, which limits its overall effectiveness.

\noindent
\textbf{Unsupervised Generalization.} Following the methodology of constructing
the Mono for Stereo (MfS) dataset in ~\cite{watson2020learning}, we construct
our training dataset, MfS-GenStereo, from the same datasets, including ADE20K~\cite{zhou2017scene},
Mapillary Vistas~\cite{neuhold2017mapillary}, DIODE~\cite{diode_dataset},
Depth in the Wild~\cite{chen2016single}, and COCO 2017~\cite{lin2014microsoft}.
During the generation process, we randomly select a scale factor in the range $[0
, 0.35]$ and set the maximum disparity to 256 for PSMNet. As shown in
\cref{tab:unsupervised_generalize}, our method enables unsupervised stereo matching
using only left images, achieving superior performance.

\begin{table*}
  [th]
  \centering
  \scriptsize
  \caption{Ablation studies. SD v1.5 is used for the experiments.}
  \begin{tabular}{c c c c c c c c|c c c|c c c}
    \toprule \multirow{2.5}{*}{\shortstack{\textbf{Disparity} \\ \textbf{Source}}} & \multirow{2.5}{*}{\#} & \multirow{2.5}{*}{\shortstack{\textbf{Mixed} \\ \textbf{Datasets}}} & \multirow{2.5}{*}{\shortstack{\textbf{Random} \\ \textbf{Drop}}} & \multirow{2.5}{*}{\shortstack{\textbf{Coord} \\ \textbf{Embedding}}} & \multirow{2.5}{*}{\shortstack{\textbf{Warped} \\ \textbf{Image}}} & \multirow{2.5}{*}{\shortstack{\textbf{Pixel} \\ \textbf{Loss}}} & \multirow{2.5}{*}{\textbf{Fusion}} & \multicolumn{3}{c|}{\textbf{Middlebury 2014}} & \multicolumn{3}{c}{\textbf{KITTI 2015}} \\
    \cmidrule(lr){9-11} \cmidrule(lr){12-14}                                       &                       &                                                                     &                                                                  &                                                                      &                                                                   &                                                                 &                                    & \textbf{PSNR}                                 & \textbf{SSIM}                          & \textbf{LPIPS}            & \textbf{PSNR}             & \textbf{SSIM}            & \textbf{LPIPS}            \\
    \midrule \multirow{2.5}{*}{\shortstack{Ground \\ Truth}}                       & (1)                   & \cmark                                                              & \xmark                                                           & \cmark                                                               & \cmark                                                            & \cmark                                                          & \cmark                             & 23.552                                        & 0.862                                  & 0.0641                    & 16.164                    & 0.664                    & 0.1740                    \\
                                                                                   & (2)                   & \cmark                                                              & \cmark                                                           & \cmark                                                               & \cmark                                                            & \cmark                                                          & \cmark                             & \capscore{gold}{23.699}                       & \capscore{gold}{0.866}                 & \capscore{gold}{0.0635}   & \capscore{gold}{20.439}   & \capscore{gold}{0.746}   & \capscore{gold}{0.1079}   \\
    \midrule \multirow{6}{*}{\shortstack{Pseudo \\ Disparities}}                   & (3)                   & \xmark                                                              & \cmark                                                           & \cmark                                                               & \cmark                                                            & \cmark                                                          & \cmark                             & 20.037                                        & 0.745                                  & 0.1214                    & 21.085                    & 0.777                    & 0.1113                    \\
                                                                                   & (4)                   & \cmark                                                              & \cmark                                                           & \cmark                                                               & \xmark                                                            & \xmark                                                          & \xmark                             & 22.033                                        & 0.796                                  & 0.0731                    & 21.866                    & 0.779                    & 0.1002                    \\
                                                                                   & (5)                   & \cmark                                                              & \cmark                                                           & \xmark                                                               & \cmark                                                            & \xmark                                                          & \xmark                             & 22.697                                        & 0.832                                  & 0.0715                    & 22.313                    & 0.798                    & 0.1011                    \\
                                                                                   & (6)                   & \cmark                                                              & \cmark                                                           & \cmark                                                               & \cmark                                                            & \xmark                                                          & \xmark                             & 23.407                                        & 0.855                                  & 0.0644                    & 22.520                    & 0.805                    & 0.0974                    \\
                                                                                   & (7)                   & \cmark                                                              & \cmark                                                           & \cmark                                                               & \cmark                                                            & \cmark                                                          & \xmark                             & \capscore{silver}{23.811}                     & \capscore{silver}{0.867}               & \capscore{silver}{0.0621} & \capscore{silver}{22.716} & \capscore{silver}{0.811} & \capscore{silver}{0.0961} \\
                                                                                   & (8)                   & \cmark                                                              & \cmark                                                           & \cmark                                                               & \cmark                                                            & \cmark                                                          & \cmark                             & \capscore{gold}{23.835}                       & \capscore{gold}{0.868}                 & \capscore{gold}{0.0620}   & \capscore{gold}{22.749}   & \capscore{gold}{0.813}   & \capscore{gold}{0.0958}   \\
    \bottomrule
  \end{tabular}
  \label{tab:ablation_studies}
\end{table*}

We compare our approach with the original MfS~\cite{watson2020learning} and its enhanced
variant using the Depth Anything Model v2 (DAMv2), which generates the right
view images by filling the occluded areas with random background patches from
the dataset. The experiments indicate that, despite the better depth prior, the synthesis
of right-view images remains a critical bottleneck in this methodology.
Additionally, we compare with Mono2Stereo, which fine-tunes SD-Inpainting for
right-view image generation, and NeRFStereo, which generates stereo images by
reconstructing the static scenes using Neural Radiance Fields (NeRF), limiting the
diversity of its datasets. Our experimental results indicate that our proposed
method consistently surpasses these existing approaches in generating high-quality
stereo pairs with pixel-level accuracy, demonstrating robust generalization capabilities
and significantly enhancing stereo matching performance across diverse datasets.

\subsection{Ablation Study.}

\noindent
\textbf{Random disparity drop.} Since the ground-truth disparity maps in real-world
datasets (e.g., KITTI) are often derived from LiDAR, resulting in sparse and
incomplete point clouds, our random disparity dropout strategy mitigates the model's
reliance on dense disparity supervision. This enhancement allows the model to
generate plausible right images even when only sparse disparity maps are
available. This strategy significantly improves performance on datasets with sparse
ground-truth annotations, demonstrating better generalization to real-world
scenarios. As shown in \cref{tab:ablation_studies}, comparing rows (1) and (2)
highlights the effectiveness of this strategy.

\noindent
\textbf{Mixed datasets.} We train our model on Virtual KITTI 2 (VKITTI2)~\cite{cabon2020virtual}
and a combination of mixed datasets, and subsequently evaluate both models. When
trained exclusively on VKITTI2, the model achieves comparable performance to the
mixed-dataset model on KITTI 2015, owing to the relatively small domain gap between
VKITTI2 and KITTI 2015. However, this single-dataset training results in notable
performance degradation on other datasets, indicating limited generalization. As
shown in \cref{tab:ablation_studies}, comparing rows (3) and (8) demonstrates that
our mixed dataset training strategy significantly improves the model's generalization
across diverse domains.

\noindent
\textbf{Coordinate embedding and the warped image.} Incorporating the coordinate
embedding process effectively encodes spatial information and reinforces the
structural relationship between stereo views. Additionally, using the warped
image as a condition enhances performance by providing pixel-level guidance. The
warped image serves as a coarse prediction of the right view, helping the model
focus on correcting local inconsistencies rather than synthesizing the entire
image from scratch. As shown in rows (4), (5), and (6) in \cref{tab:ablation_studies},
the combination of coordinate embedding and warped image significantly improves
generation performance, preserving geometric consistency and reducing artifacts like
stitching boundaries. While warped image embedding contributes most to
performance, its effectiveness decreases with additional fine-tuning iterations,
revealing instability when using warped image alone as a conditioning signal. This
underscores the importance of combining coordinate embedding and warped image to
achieve stable, high-quality stereo image generation.

\noindent
\textbf{Pixel-level loss.} By decoding the latent representations back to the image
space, the model learns to minimize discrepancies directly in the observed image
space, enforcing pixel-level alignment. As illustrated in rows (6) and (7), the introduction
of the pixel-level loss significantly improves the generation performance,
highlighting its role in maintaining high-fidelity stereo image generation.

\noindent
\textbf{Adaptive Fusion.} The proposed adaptive fusion module significantly enhances
visual quality by dynamically weighting the contributions of both the generated and
warped images. A comparison of rows (7) and (8) demonstrates that this fusion
strategy further refines pixel alignment, resulting in sharper, more
structurally coherent, and perceptually accurate reconstructions. The
improvement in PSNR is more pronounced than that in SSIM and LPIPS, indicating that
the adaptive fusion is particularly effective for pixel-level alignment.
\section{Conclusion}
We presented GenStereo, a novel diffusion-based framework for open-world
stereo image generation with applications in unsupervised stereo matching. Our method
introduces several key innovations, including disparity-aware coordinate
embeddings along with warped image embeddings, pixel-level loss, and an adaptive
fusion module, to ensure high-quality stereo image generation with strong geometric
and semantic consistency. Through extensive experiments across diverse datasets,
we show that GenStereo significantly outperforms existing methods in both
stereo generation and unsupervised matching. Our ablation studies demonstrate
the efficacy of each proposed component.

\noindent
\textbf{Limitations.} Despite promising results, diffusion-based models face
inherent challenges with large disparities when generating right-view images due
to large unconditioned regions. While our data augmentation such as random cropping
and resizing mitigates these issues and performs well in typical stereo setups,
future work could explore extending these models to accommodate larger
disparities.

\clearpage

\section*{Acknowledgments}
We gratefully acknowledge the advanced computational resources provided by Engineering IT and Research Infrastructure Services at Washington University in St. Louis.

{
    \small
    \bibliographystyle{ieeenat_fullname}
    \bibliography{main}
}


\end{document}